\documentclass[10pt,journal]{IEEEtran}
\usepackage[pdftex]{graphicx}
\DeclareGraphicsExtensions{.pdf,.jpg,.png}
\usepackage{cite}
\usepackage[cmex10]{amsmath}
\usepackage{amssymb}
\usepackage{graphicx}
\usepackage{cases}
\usepackage{mdwmath}
\usepackage{mdwtab}
\usepackage{array}
\usepackage{url}
\usepackage[ruled,vlined]{algorithm2e}
\usepackage{booktabs}
\usepackage{threeparttable}
\usepackage{multirow}
\usepackage{color}
\usepackage{caption}

\def\eg{{\em e.g.}}
\def\etal{{\em et al.}}

\newcommand{\myPara}[1]{\vspace{.05in}\noindent\textbf{#1}}

\newcommand{\bl}[1]{\textbf{#1}}

\newcommand{\mc}[1]{\mathcal{#1}}

\newcommand{\bm}[1]{\mbox{\boldmath{$#1$}}}

\graphicspath{{./fig/}}
\usepackage{color}
\begin{document}

\title{Distilling Channels for Efficient Deep Tracking}
\author{Shiming Ge,~\IEEEmembership{Senior Member,~IEEE,}
    Zhao Luo, Chunhui Zhang, Yingying Hua and
    Dacheng~Tao,~\IEEEmembership{Fellow,~IEEE}% <-this % stops a space
\thanks{S. Ge is with the Institute of Information Engineering, Chinese Academy of Sciences, Beijing, 100095, China. e-mail: geshiming@iie.ac.cn.}
\thanks{Z. Luo, C. Zhang and Y. Hua are with the Institute of Information Engineering, Chinese Academy of Sciences, and School of Cyber Security at University of Chinese Academy of Sciences.}
\thanks{D. Tao is with the School of Information Technologies and the Engineering and Information Technologies in the University of Sydney. e-mail: dacheng.tao@sydney.edu.au }% <-this % stops a space
}

\markboth{Submission to IEEE Transactions on Image Processing}%
{Ge \MakeLowercase{\textit{et al.}}: Bare Demo of IEEEtran.cls for Journals}

\maketitle

\begin{abstract}
  Deep trackers have proven success in visual tracking. Typically, these trackers employ optimally pre-trained deep networks to represent all diverse objects with multi-channel features from some fixed layers. The deep networks employed are usually trained to extract rich knowledge from massive data used in object classification and so they are capable to represent generic objects very well. However, these networks are too complex to represent a specific moving object, leading to poor generalization as well as high computational and memory costs. This paper presents a novel and general framework termed channel distillation to facilitate deep trackers. To validate the effectiveness of channel distillation, we take discriminative correlation filter (DCF) and ECO for example. We demonstrate that an integrated formulation can turn feature compression, response map generation, and model update into a unified energy minimization problem to adaptively select informative feature channels that improve the efficacy of tracking moving objects on the fly. Channel distillation can accurately extract good channels, alleviating the influence of noisy channels and generally reducing the number of channels, as well as adaptively generalizing to different channels and networks. The resulting deep tracker is accurate, fast, and has low memory requirements. Extensive experimental evaluations on popular benchmarks clearly demonstrate the effectiveness and generalizability of our framework.
\end{abstract}

% like video surveillance and autonomous driving


\IEEEpeerreviewmaketitle

\section{Introduction}

Efficient visual tracking is important in many real-world computer vision and multimedia applications including video analysis~\cite{Zhang2017MM,feichtenhofer2017iccv}, video surveillance~\cite{Smeulders2014PAMI}, automatic pilotings~\cite{Galoogahi2017ICCV} and human computer interactions~\cite{yang2017deep}. Recently, state-of-the-art visual trackers~\cite{Ma2015ICCV,Qi2016CVPR,Nam2016CVPR,Danelljan2016ECCV,Danelljan2017CVPR,Song2017ICCV}, which are usually based on deep networks, have performed extremely well in various popular benchmarks. For example, by employing a deep network VGG-M~\cite{Simonyan2015ICLR} trained for object classification on ImageNet~\cite{Russakovsky2015IJCV} with a top-5 accuracy of $84.1\%$, the ECO tracker~\cite{Danelljan2017CVPR} achieved a precision of $91\%$ on the OTB100 benchmark~\cite{Wu2015PAMI}, and the C-COT tracker~\cite{Danelljan2016ECCV} delivered a high expected average overlap (EAO) of $0.281$ on the VOT2017 benchmark~\cite{Kristan2017a}.

Although the superior representation power of deep networks results in high accuracy, these trackers usually incur  high computational and memory costs, which decreases tracking efficacy and hinders their practical deployment on resource-limited devices. These huge costs arise from the redundancy in representing specific tracked objects using deep networks for generic objects.
Some proposed direct solutions~\cite{Nam2016CVPR,Guo2017ICCV,Song2017ICCV} apply online model update to adapt deep networks learned for object classification or detection for tracking.
Although they perform well, these approaches are expensive and inefficient. It is therefore necessary to develop an efficient deep tracker whilst preserving accuracy. To improve viusal tracking efficacy, a number of deep trackers have been proposed that can be grouped into three main categories according to their feature processing scheme: learning, weighting, and compression trackers.

``Learning'' deep trackers directly learn new compact deep networks for feature representation in tracking from massive annotated visual data. For example, Bertinetto~\etal~\cite{Bertinetto2016ECCVW} proposed tracking objects by offline training a fully-convolutional Siamese network on the ILSVRC15 dataset for object detection in videos. Compared to the deep networks learned for object classification, the trained Siamese network had a more adaptive expressive power for various objects. Valmadre~\etal~\cite{Valmadre2017CVPR} proposed turning correlation filters into a differentiable layer in a deep neural network, and then learning deep features end-to-end, tightly coupled to the correlation filter. In general, the learning-based deep trackers need extra large-scale training from massive data, and the most important component in these approaches is transferring the knowledge learned from various objects to a specific object during tracking. However, a key issue must be carefully addressed in these approaches is how to adaptively transfer the desired rather than total knowledge from learned objects.

``Weighting'' deep trackers weight deep features or responses from different layers of a deep network pre-trained on object classification by adaptively measuring layer influences. For example, Ma \etal~\cite{Ma2015ICCV} observed that the earlier layers provide more precise localization, while the later convolutional layers encode the semantic information of objects. They then proposed to hierarchically exploit both facets by fusing their confidence responses. In contrast, Qi~\etal~\cite{Qi2016CVPR} presented an adaptive hedge method to combine features from different deep layers into a multi-channel feature containing a large channel number (3,072). Some deep trackers performed weighting by using attention mechanism~\cite{Cui2016CVPR}, spatial reliability~\cite{Lukezi2017CVPR}, reinforcement learning~\cite{Yun2017CVPR} or multiple templates~\cite{Liu2018TIP}. Typically, these trackers have impressive accuracy when tracking very different objects, but the feature channel number remains huge.

``Compression'' deep trackers reduce or compress the feature dimension. Danelljan~\etal~\cite{Danelljan2017PAMI} used classical dimension reduction such as with principal component analysis (PCA) to reduce multi-channel features. Later, they~\cite{Danelljan2017CVPR} proposed factorized convolution to accelerate the main convolution computation in deep network inference. Choi~\etal~\cite{Choi2018CVPR} proposed deep feature compression for fast tracking by a context-aware scheme utilizing multiple expert auto-encoders. These compression-based trackers mainly aim to reduce the number of network parameters, so do not overcome the high memory complexity required for deep feature extraction.

In summary, the multi-channel features describing an object in different views used in most deep trackers are fixed for diverse tracked objects. For visual tracking, the core problems are (i) how to adaptively distill the correct knowledge from general knowledge learned from massive data, and (ii) how to transfer it to track a specific object on the fly. The fixed feature channel setting degrades tracking performance. In this paper, we investigate the effect of channel selection on tracking performance and show that precision and success rates can be significantly improved by selecting informative channels and removing noisy channels. Based on this finding, we propose a novel channel selection framework called ``channel distillation'' to learn a best binary weighting to compress features. To validate the effectiveness of channel distillation, we take standard DCF as an example and integrate them. In this way, feature compression, response map generation, and model update are optimized in a unified and integrated formulation. We show through extensive evaluations on videos from popular benchmarks that integrating our channel distillation framework improves the tested deep trackers.

Our main contributions are as follows. First, we investigate the influence of channel selection for multi-channel features on tracking performance and reveal that there exist specific optimal channels for diverse tracked objects. Second, we propose channel distillation to adaptively select good channels and, by way of illustration, use DCF to formulate the integrated framework as an energy minimization problem, thereby improving accuracy, speed, and memory storage. Finally, we conduct a comprehensive evaluation and analysis to show the effectiveness and generalizability of channel distillation, which may be helpful for developing efficient deep trackers for real-world applications.

\section{Related Work}
\subsection{Multi-channel Deep Features}
The types of features employed in visual trackers significantly affect tracking performance. Encouraged by deep learning~\cite{Alex2012NIPS}, recent visual trackers mainly apply deep features instead of the single or multi-channel hand-crafted features used in early DCF trackers~\cite{Bolme2010CVPR,Henriques2015PAMI}.

Ma~\etal~\cite{Ma2015ICCV} employed multiple convolutional layers to improve tracking accuracy by hierarchically utilizing deep features from the early and final layers in the DCF framework. Danelljan~\etal~\cite{Danelljan2016ECCV} used $611$-channel multi-resolution deep feature maps in a continuous formulation to improve tracking performance. Generally, these deep trackers performed significantly better and gained higher accuracy than other trackers using hand-crafted features. Some methods prefer to mix or combine features to improve them. Qi~\etal~\cite{Qi2016CVPR} used an adaptive hedge method to combine features from different convolutional layers into a single layer. He~\etal~\cite{He2018CVPR} adopted a channel attention mechanism to weight different channels. {Wang~\etal~\cite{RASNet} presented a residual attentional Siamese network to reformulate the correlation filter within a Siamese tracking framework, and introduces different kinds of the attention mechanisms to adapt the model without updating the model online. Song~\etal~\cite{Song2017ICCV} proposed reformulating discriminative correlation filter (DCF) as a one-layer convolutional neural network, using VGG-16 as the feature extractor.} In~\cite{He2017ICCVW}, the authors weighted convolution responses from each feature block and then summed them to produce the final confidence score. Huang~\etal~\cite{Huang2017ICCV} proposed an approach to improve deep tracker speed by adaptively processing easy frames with cheap pixel features and challenging frames with expensive deep features. {Wang~\etal~\cite{wang2015ICCV} proposed two networks to online select feature maps from different layers of VGG-16. Danelljan~\etal~\cite{Danelljan2016ECCV} used multi-resolution deep feature maps in a continuous formulation. Lu~\etal~\cite{DSLT} applied residual connections to fuse multiple convolutional layers as well as their output response maps.} Choi~\etal~\cite{Choi2017CVPR} introduced a deep attentional network to choose a subset of associated correlation filters according to tracked object  s dynamic properties. Typically, these methods improved tracking accuracy, but the mixed approach using weighting or combination could not reduce the computation and storage requirements of original features.

In summary, deep features from well pre-trained networks are usually sufficient to represent generic objects, and fully fixed feature channels are typically used by current deep trackers. However, these deep features usually contain significant redundancy, and the features used in fixed channels generally incur large memory and computational costs. It is therefore necessary to reduce this redundancy to improve tracking efficacy.

% framework
\begin{figure*}[t]
  \centering{\includegraphics[width=1.0\linewidth]{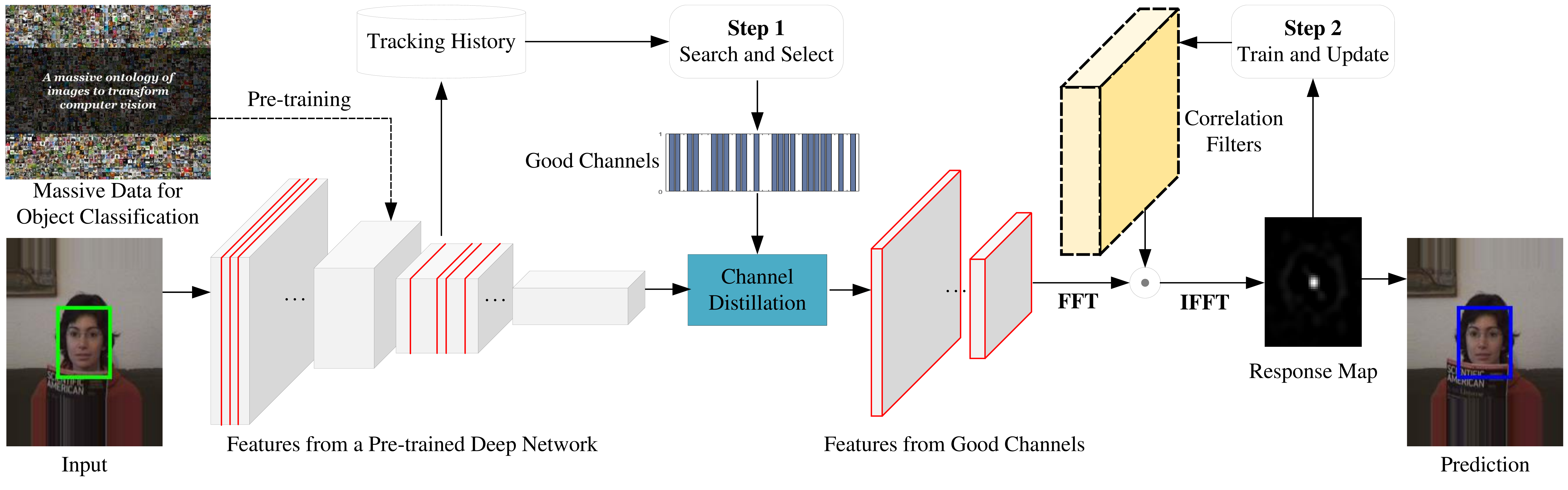}}
  \caption{ Schematic of our tracking framework with channel distillation. Channel distillation adaptively selects good channels from a deep network pre-trained for object classification to represent diverse tracked objects, then the distilled features are fed to DCF or ECO, forming an integrated deep tracking formulation. The formulation addresses feature compression, response map generation and model update in a unified framework.}
\label{fig:framework}
\end{figure*}

\subsection{Feature Compression Methods}
Feature compression represents an alternative way to improve tracking efficacy. Feature compression includes channel pruning and dimension reduction.

\myPara{Channel Pruning.}~Noting that a multi-channel feature can describe an object from various views with different channels, a given object may have some specific channel features. Therefore, several methods have been proposed to prune channels to reduce the feature representation. Channel-pruning methods have recently been used to remove redundant channels in the feature maps of deep convolutional neural networks. In this way, the trained deep models could be condensed and the inference time reduced. To accelerate very deep models, He~\etal~\cite{He2017ICCV} introduced an iterative, two-step algorithm to prune each layer by LASSO regression-based channel selection and least squares reconstruction. The method achieved a two- to five-times increase in speed with very small accuracy loss for object classification. Similarly, Liu~\etal~\cite{Liu2017ICCV} proposed a network slimming approach to enforce channel-level sparsity in the network by automatically identifying and pruning insignificant channels. Wang ~\etal~\cite{Wang2017ICME} proposed to transfer features for object classification to tracking domain via convolutional channel reductions. They viewed channel reduction as an additional convolutional layer with a specific task. This approach not only extracted useful information for tracking performance but also significantly increased tracking speed. Some model compression methods with knowledge distillation~\cite{Hinton2015Distilling,romero2015fitnets} modify deep networks to improve efficacy.

\myPara{Dimension Reduction.} Beyond channel pruning, some recent works have attempted to compress feature dimensions to improve speed and reduce the memory footprint. Danelljan~\etal~\cite{Danelljan2017PAMI} proposed fDSST to compress HOG features with PCA to achieve $54.3$ fps using a single CPU. Later, they proposed ECO~\cite{Danelljan2017CVPR}, which employed the factorized convolution operator to compress deep features, and this operated at 8 fps using a high-end GPU. Xu and Lu~\cite{Xu2017MMM} reported a multi-channel compressive feature to describe objects. Their method combined rich information from multiple channels and then projected it into a low-dimension compressive feature space. {Gundogdu and Alatan~\cite{CFCF} proposed fine-tuning the convolutional parts of a state-of-the-art deep network and integrating this model to a correlation filter-based tracker. Chen and Tao~\cite{CRT} proposed learning a regression model for visual tracking with single convolutional layer.} Choi~\etal~\cite{Choi2018CVPR} proposed a context-aware scheme for feature compression by utilizing multiple expert auto-encoders.
Generally, most of these feature compression methods focus on pruning or compressing deep features in object classification. Due to the computational expense of deep feature extraction, these trackers still have a high memory cost.

Inspired by this and in contrast to approaches that improve multi-channel feature representation by weighting or reducing deep features, here we adaptively select representative channels from deep features by energy minimization, which boosts both tracking accuracy and speed whilst consuming less memory. Our channel distillation framework is a general feature channel selection method that could easily be integrated into other frameworks to improve tracking performance of diverse objects in different videos. Taking DCF as an example, we integrate it into a unified formulation that simultaneously addresses feature compression, response map generation and model update.

\section{Our Approach} \label{sec:approach}
In this section, we first review the general channel distillation formulation. We then study the effect of channel pruning and selection on tracking performance by conducting experiments that demonstrate the existence of good channels for tracking a specific object. Based on this finding, we propose channel distillation and formulate tracking as an energy minimization problem by incorporating it into the DCF and ECO framework (see Fig.\ref{fig:framework}). Finally, we propose an alternating optimization algorithm to solve this problem.

\subsection{Channel Distillation Formulation} \label{subsection:cd}
Channel distillation aims to extract informative channels and prune noisy channels by adaptively selecting the best channels for diverse tracked objects, which makes the distilled feature channels powerful for improving tracking performance.

The objective of channel distillation is to learn both tracking model $\bl{h}$ and good channel selection $\bl{a}$ from a set of training examples $\{(\bl{x}_i,\bl{y}_i)\}_{i=1}^n$ by using a multi-channel deep feature descriptor $\bl{f}=\{f^{(l)}\}_{l=1}^{d}$, where $d\ge1$ and $n$ are the numbers of feature channels and training examples respectively, $\bl{x}_i$ is the authentic input image and $\bl{y}_i$ is its ideal desired output or response map. $\bl{x}_i$ is represented as a $d$-channel feature with $\bl{f}$. Therefore, the channel distillation formulation aims to minimize the following loss $E$:
\begin{equation}
E(\bl{h},\bl{a}) = \sum_{i=1}^{n}{\mc{L}\left(\sum_{l=1}^{d}{\alpha_l} {\phi}\left(f^{(l)}\left(\bl{x}_i\right), \bl{h}\right),\bl{y}_i\right)} + {\lambda} \mc{R}(\bl{h}),
\label{eq:cd}
\end{equation}
where $\mc{L}(\cdot)$ is a function to measure the difference between predicted and ideal desired output, ${\phi}(\cdot)$ is the model matching operator, $\mc{R}(\cdot)$ is a regularization function, and parameter $\lambda\ge0$ is used to balance the two energy terms. $\bl{a}=(\alpha_1,...\alpha_l,...\alpha_d)$ is a $d$-dimensional binary vector for encoding the channel selection, and $\alpha_l$ indicates whether the $l$-th channel is selected~($\alpha_l=1$) or pruned~($\alpha_l=0$).

It can be seen that solving Eq.\eqref{eq:cd} means to learn a binary weighting to achieve a compressed feature, so channel distillation unifies three tracker categories. It is also noted that channel distillation is different from traditional three categories including: i) channel distillation uses binary rather than real-value weights to encode channel weighting, which greatly saves memory storage for feature maps and tracking models, ii) channel distillation provides an economic and feasible way to re-use pre-trained deep networks, avoiding large-scale learning a new one from massive data, and iii) channel distillation can compress the dimension of deep features to reduce both memory and model storage, rather than only model size reduced in some compression-based deep trackers.

It also can easily be seen that the channel distillation in Eq.\eqref{eq:cd} is generalizable to other tracking formulations. For instance, in the fully-connected Siamese network formulation~\cite{Bertinetto2016ECCVW}, the features are extracted from the input image as a large search area while the tracking model $\bl{h}$ is for the object. Then, $\bl{y}_i$ is a response map in the search area that represents the object location.
The deep regression formulation~\cite{Held2016ECCV} has similar tracking model, when $\bl{y}_i$ is a direct location output represented as a $2$-dimensional vector.
In various DCF formulations \cite{Bolme2010CVPR,Galoogahi2013ICCV,Danelljan2015ICCV,Danelljan2015ICCVW,Mueller2017CVPR}, $\bl{h}$ is a multi-channel spatial frequency array (equivalently, a template in image domain) with the same dimension as multi-channel features and $\bl{y}_i$ is a response map with the same width and height as the tracking input image. The channel candidates for selection are also flexible, implying that the multi-channel features can be selected from one or multiple deep networks, hand-crafted features~\cite{Huang2017ICCV}, or their mixtures (\eg, cascaded cheap and deep features in \cite{Huang2017ICCV}).
%Channel distillation can evaluate multiple feature channels and select the most informative ones with respective to energy minimization.

\subsection{Existence of Good Channels} \label{subsec:csa}
\begin{figure*}[t]
\centering\centerline{\includegraphics[width=1.0\linewidth]{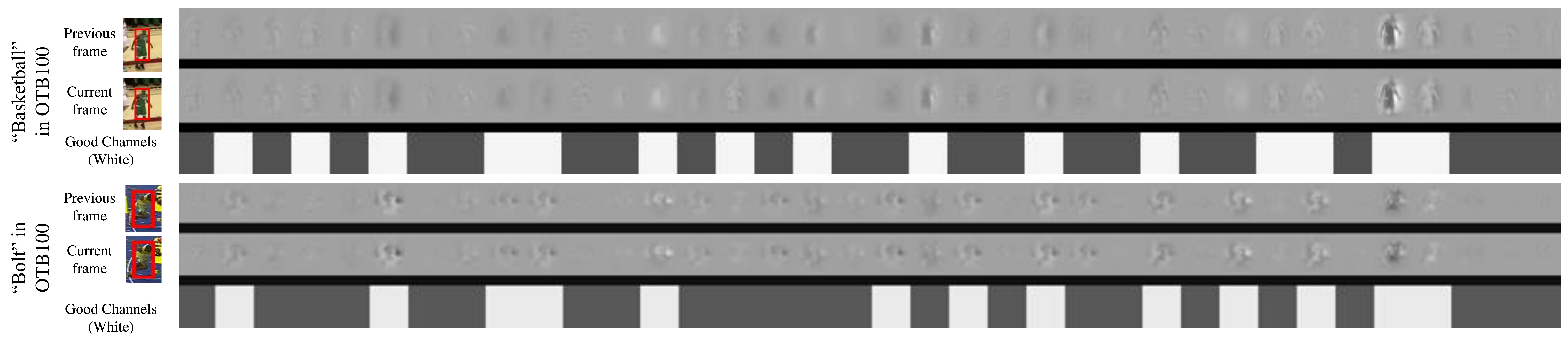}}
\caption{Examples of ``good'' channels in two video sequences selected from a convolutional layer. The 1st and 2nd rows visualize the multi-channel features in two consecutive frames. The 3rd row shows the selected good channels in white, which are spatially salient and temporally consistent. This implies that good channels for a specific tracked object exist in a video.}
\label{fig:fe_analysis}
\end{figure*}

Eq.\eqref{eq:cd} shows that good channels are evaluated and selected according to energy minimization. We take DCF for example to study their existence. Considering that the response map is generated by accumulating multiple channel-wise correlation outputs, we first experimentally analyze the influence of channel selection (or pruning) on tracking performance by investigating the typical DCF tracker~\cite{Ma2015ICCV} with the VGG-19 model on ImageNet~\cite{Simonyan2015ICLR}. This analysis is readily extendable to other DCF frameworks (e.g., CACF~\cite{Mueller2017CVPR}) and multi-channel features from other deep networks. The objective of this analysis is to simply reveal the $\textbf{existence}$ of channel selectivity and provide insights to guide the design of our framework. To this end, we analyze good channels by supposing that the object position in the current frame is known (e.g., by the ground truth).

For an object $\bl{o}$ to be tracked in a video $\bl{v}$ containing $k$ frames, feature channel selection aims to find the optimal feature channels $\bl{C}=\{C_i\}_{i=1}^c$ achieving the best tracking performance, where $c\le d$ and $C_i\in\{1,2,...,d\}$. In our experiments, we perform channel selection based on the empirical consideration that the selected feature channels are more discriminative in spatial and stable in temporal when tracking the object. {Here, spatial discrimination means the feature channel is more salient or ``good'' to measure the target features such that the object can be identified from background distractors, while temporal stability indicates that the target features in two consecutive frames are consistent such that the object can be robustly tracked. We call the selected feature channels for an object $\bl{o}$ in a video $\bl{v}$ as its ``good channels'', and represent the priority of a channel as its ``friendliness''. The friendliness of a channel reflects its contribution to the tracking performance.} To this end, we select feature channels via the following four stages:

1)~The object patch $\bl{x}_i$ in a video is cropped from the $i$-th image frame $\bl{F}_i$ according to its ground truth bounding box $\bl{b}_i$, and then fed to a pre-trained VGG-19 to generate a multi-channel feature $\bl{f}_i=\{f_i^{(l)}\}_{l=1}^{d}$.

2)~The spatial discrimination and temporal stability in the $l$-th channel are calculated between the objects in two consecutive frames as $s_i(l)=\|f_i^{(l)}\|/m$ and $t_i(l)=-\|f_i^{(l)}-f_{i+1}^{(l)}\|^2$ respectively, where $m$ is the number of channel elements. This shows that channels with larger $s_i(l)$ and $t_i(l)$ tend to be good channels. Here, we generally consider good spatial discrimination as salient feature channels with larger activations since they can produce more helpful impacts in correlation filtering, and thus measure spatial discrimination with the magnitude of the feature vector.

3)~The channel friendliness is calculated by combining $s_i(l)$ and $t_i(l)$ and is represented as $r_i(l)=(s_i(l)-1)t_i(l)$, where a larger $r_i(l)$ implies good channels give preference to the $l$-th channel in tracking the $i$-th frame. Then, the total channel friendliness is summed in all consecutive $k-1$ frames to obtain the average channel friendliness set $\{\overline{r}(l)\}_{l=1}^d$, where $\overline{r}(l)={\sum_{i=1}^{k-1}{r_i(l)}}/(k-1)$.

4)~The feature channels are ranked according to the average channel friendliness in descending order. Then, the tracking performance is evaluated by iteratively pruning the channel with the smallest average channel friendliness until the performance decreases or the maximum iteration is reached. Finally, good channels are returned as the channel set with the best tracking performance.

After selection, we study its effectiveness. Compared with the original fixed channel setting on all channel features in some layers, the total tracking precisions with good channels are improved for all videos. {The main reason comes from the redundancy due to the fundamental inconsistency between predicting object class labels in classification and locating targets of arbitrary classes in tracking (as stated in~\cite{Nam2016CVPR}), such that channel distillation in Eq.\eqref{eq:cd} can avoid learning those correlation filters that contains negligible energy. This can lead to improved tracking accuracy (as stated in~\cite{Danelljan2017CVPR}).}

Moreover, we also find that good channels mainly focus on the channels in which the features are often more spatially salient and temporally consistent. An example is shown in Fig.\ref{fig:fe_analysis}. These findings imply that: 1) there exists a specific good channel set for a tracked object in a video; 2) some noisy channels should be discarded or pruned to improve performance; and 3) the same types of objects (e.g., human) share some similar good channels, indicating that channel selectivity may arise from similar variations in specific objects with respect to appearance, motion, etc.

In summary, there exist good channels for tracking in a video that are useful for improving tracking performance. We next examine how to select these ``good channels'' for tracking.

\subsection{Channel Distillation in DCF} \label{subsection:dcf}
Despite the existence of good channels, selecting them is challenging since the object in the tracking frames is unknown in advance except for in the first frame. We therefore apply channel distillation to address this issue. Due to its popularity and efficacy in visual tracking, we first use standard DCF as an example and incorporate it into channel distillation, which can be formulated as a joint optimization problem:
\begin{equation}
\begin{aligned}
E(\bl{h},\bl{a}) = \sum_{i=1}^{n}{\left\|\sum_{l=1}^{d}{\alpha_l}(f^{(l)}(\bl{x}_i)\otimes h^{(l)})-\bl{y}_i\right\|^2} \\
+\frac{\lambda}{\left\|{\bl{a}}\right\|} \sum_{l=1}^{d}{{\alpha_l}\left\|h^{(l)}\right\|^2},
~~~\text{s.t.}~~\alpha_l \in {\{0,1\}},
\end{aligned}
\label{eq:dcf-cd}
\end{equation}
where $\otimes$ denotes circular convolution, $\left\|\bl{a}\right\|$ represents the number of good channels, and the tracking model $\bl{h}=\{h^{(l)}\}_{l=1}^{d}$ is a multi-channel correlation filter. The first term is used to measure the filtering cost between the cross-correlation output and the ideal desired correlation output for authentic input images, while the second term is for regularizing the correlation filter. Based on Parseval's formula, the problem can be transformed into a frequency domain form. Denote the discrete Fourier transform~(DFT) operator as $\hat{f}=\mathcal{F}(f)$, then Eq.\eqref{eq:dcf-cd} is rewritten as:
\begin{equation}
\begin{aligned}
E(\hat{\bl{h}},\bl{a}) = \sum_{i=1}^{n}{\left\|\sum_{l=1}^{d}{\alpha_l}(\hat{f}^{(l)}(\bl{x}_i)\odot\hat{h}^{(l)*})-\hat{\bl{y}}_i\right\|^2} \\ +\frac{\lambda}{\left\|{\bl{a}}\right\|} \sum_{l=1}^{d}{{\alpha_l}\left\|\hat{h}^{(l)}\right\|^2},~~~\text{s.t.}~~\alpha_l \in {\{0,1\}},
\end{aligned}
\label{eq:dcf-cd-fft}
\end{equation}
where $\odot$ denotes the element-wise product, and $*$ is the conjugation operator. Note that it is difficult to resolve Eq.\eqref{eq:dcf-cd} or Eq.\eqref{eq:dcf-cd-fft}, which we address with alternating optimization:

\myPara{Step 1.} In this step, the objective is to search for the optimal setting of $\bl{a}$ to minimize the loss defined in Eq.\eqref{eq:dcf-cd-fft} when giving $\bl{h}$ or $\hat{\bl{h}}$. Noting that $\bl{a}$ is a vector with discrete binary values and contained in both the numerator and denominator of the second term (which is not linear), there is no analytic solution for Eq.\eqref{eq:dcf-cd-fft}. Further, exhaustive searching is very time-consuming and impractical. We address this problem via heuristic searching as follows:

1)~Inspired by channel selectivity analysis, we first evaluate the tracking history~(historical predictions up to the current frame) and obtain the initial good channels $\bl{C}^{(0)}$ encoded as a $d$-dimensional binary vector $\bl{a}^{(0)}$.

2)~Then, we fix the number of good channels as $c=\left\|{\bl{a}^{(0)}}\right\|$ and the optimization problem can be transformed into:
\begin{equation}
\begin{aligned}
E(\bl{a}) = \sum_{i=1}^{n}{\left\|\sum_{l=1}^{d}{\alpha_l}A_{il}-B_i\right\|^2}
+\frac{\lambda}{c}\sum_{l=1}^{d}{{\alpha_l}{\gamma_l}}, \\~~~\text{s.t.}~~\alpha_l \in {\{0,1\}},
\end{aligned}
\label{eq:dcf-cd-fft-estep}
\end{equation}
where $\gamma_l=\left\|\hat{h}^{(l)}\right\|^2$ is a scalar, and $A_{il}=vec\{\hat{f}^{(l)}(\bl{x}_i)\odot\hat{h}^{(l)*}\}$ and $B_i=vec\{\hat{\bl{y}}_i\}$ are two $m$-dimensional vectors. Here, $vec(\cdot)$ is an operator to transform a matrix into a vector.

3)~With the seed $\bl{a}^{(0)}$, we perform iterative searching to find the optimal setting for minimizing Eq.\eqref{eq:dcf-cd-fft-estep}. Noting that $\bl{C}^{(0)}=\{C^{(0)}_i\}_{i=1}^{c}$ is ranked and its complement is denoted $\overline{\bl{C}}^{(0)}=\{\overline{C}^{(0)}_j\}_{j=1}^{(d-c)}$,
we start from the last element $C^{(0)}_c$ of $\bl{C}^{(0)}$, swapping it with the element in $\overline{\bl{C}}^{(0)}$ one by one to evaluate Eq.\eqref{eq:dcf-cd-fft-estep} with the new good channel setting $\{C^{(0)}_i\}_{i=1}^{c-1}\bigcup\{\overline{C}^{(0)}_j\}$. If the loss decreases most and lower than the current setting $\bl{C}^{(0)}$, then we discard $C^{(0)}_c$ and update good channels with $\bl{C}^{(1)}=\{C^{(0)}_i\}_{i=1}^{c-1}\bigcup\{\overline{C}^{(0)}_j\}$. The iteration continues for all the elements in $\bl{C}^{(0)}$, and generates the final good channels $\bl{C}$ which can be encoded into the binary channel selection vector $\bl{a}^{(t)}$.

\myPara{Step 2.} In this step, given the channel selection vector $\bl{a}^{(t)}$, the correlation filter $\bl{h}$ or $\hat{\bl{h}}$ can be efficiently solved by standard DCF in the Fourier domain. Denote $\hat{f}_{ij}=\hat{f}^{C_j}(\bl{x}_i)$, $\hat{h}_j=\hat{h}^{C_j}$ and good channels $\bl{C}=\{C_j\}_{j=1}^{c}$ where $C_j\in \{1,2,..,d\}$. Then Eq.\eqref{eq:dcf-cd-fft} is rewritten as:
\begin{equation}
\begin{aligned}
E(\hat{\bl{h}}_r)= \sum_{i=1}^{n}{\left\|\sum_{j=1}^{c}{\hat{f}_{ij}}\odot{\hat{h}_j^{*}}-\hat{\bl{y}}_i\right\|^2}+\frac{\lambda}{c}\sum_{j=1}^{c}{\left\|\hat{h}_j\right\|^2},
\end{aligned}
\label{eq:dcf-cd-fft-mstep}
\end{equation}
where $\hat{\bl{h}}_r=[{\hat{h}_1^{T}},{\hat{h}_2^{T}},...,{\hat{h}_c^{T}}]$ is a super vector of the DFTs of each good channel. Suppose that
\begin{equation}
\begin{aligned}
\hat{\bl{f}_i}=[diag(\hat{f}_{i1})^{T},diag(\hat{f}_{i2})^{T},...,diag(\hat{f}_{ic})^{T}],
\end{aligned}
\label{eq:dcf-cd-fft-solution}
\end{equation}
where $diag(\cdot)$ is the operator that transforms a vector into a diagonal matrix, then the solution can be achieved with~\cite{Galoogahi2013ICCV}:
\begin{equation}
\begin{aligned}
{\hat{\bl{h}}^{*}_r} =\left(\frac{\lambda}{c}\bl{I}+\sum_{i=1}^{n}{\hat{\bl{f}_i}}^{T}{\hat{\bl{f}}_i}\right) ^{-1}\sum_{i=1}^{n}{\hat{\bl{f}_i}}^{T}{\hat{\bl{y}}_i},
\end{aligned}
\label{eq:dcf-cd-fft-solution}
\end{equation}
where $\bl{I}$ is the identity matrix. As stated in~\cite{Galoogahi2013ICCV}, $\hat{\bl{f}}_i$ is sparse banded and the efficient solution can be found by solving $m$ independent $c\times c$ linear systems, where $m$ is the signal length~(the number of channel elements). This results in a small computational cost of $O(mc^3+nmc^2)$ and memory cost of $O(mc^2)$~\cite{Galoogahi2013ICCV}. Finally, with the learned correlation filter, patterns of interest in images are searched by cross correlating the input image $\bl{x}$ and generating a response map $\bl{y}_r$:
\begin{equation}
\begin{aligned}
\bl{y}_r= \mathcal{F}^{-1}\left({\sum_{j=1}^{c}{\hat{f}_{j}}\odot{\hat{h}_j^{*}}}\right),
\end{aligned}
\label{eq:response-cd}
\end{equation}
where $\mathcal{F}^{-1}(\cdot)$ is the inverse DFT operator, and $\hat{f}_{j}=\hat{f}^{C_j}(\bl{x})$. By examining the resulting $\bl{y}_r$ for possible correlation peaks, the position of the tracked object can be determined with:
\begin{equation}
\tilde{p}=\mathop{\arg\max}_{p} {\{\mathrm{\bl{y}_r} (p)\}},
\label{eq:responsedicision}
\end{equation}
where $p=(x,y)$ denotes a coordinate position offset.

The algorithm iterates until the given maximum number is reach or the loss does not alter. In the initialization step, we perform channel selection in just two frames, which is still effective in our experiments.

It can be seen that channel distillation saves memory since it is not necessary to store all features in the last layer which typically contains more channels than early layers, while the reduced channels further reduce memory for feature computation and matching in DCF.

\subsection{Channel Distillation in ECO}
\begin{figure*}[t]
  \centering{\includegraphics[width=1.0\linewidth]{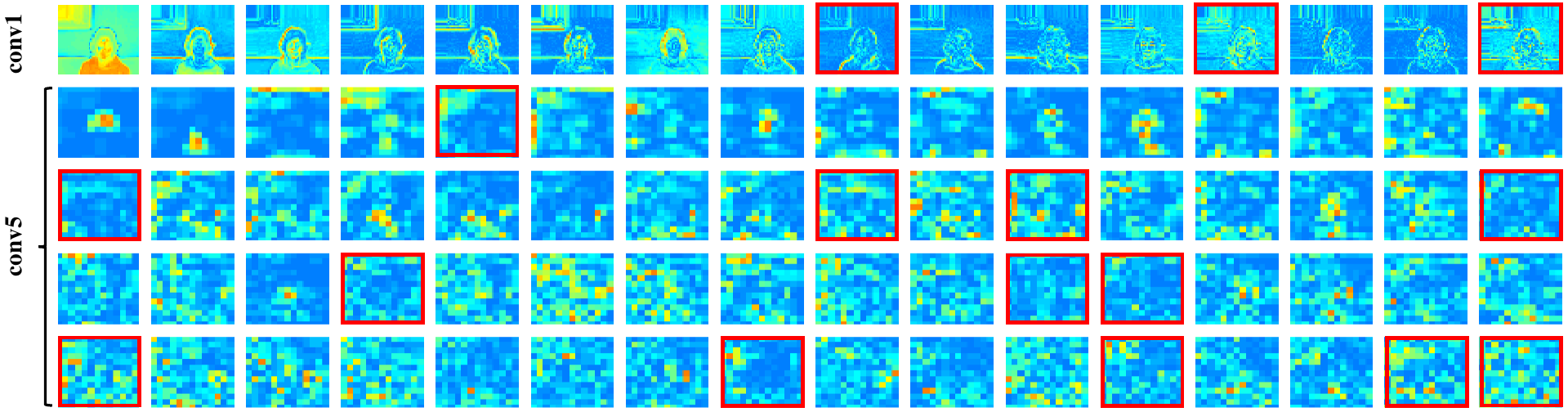}}
  \caption{An example of visualizing the feature maps when integrating channel distillation into ECO. The pruned feature channels are marked with red rectangles.}
\label{fig:visualize-eco-cd}
\end{figure*}

In this section, we take ECO~\cite{Danelljan2017CVPR} as another representative example and integrate it with channel distillation to demonstrate the generalizability of the proposed framework. ECO applies a spatial regularized variant of DCF termed SRDCF to achieve very impressive performance on recent popular benchmarks. Inspired by that many filters containing negligible energy produce unhelpful feature channels, ECO proposed using a factorized convolution operator to reduce the number of filters or parameters in the tracking model with a learned projection matrix after extracting deep features. To this end, ECO is formulated as the minimal of the loss function
\begin{equation}
\begin{aligned}
E(\bl{h},\bl{P}) = \sum_{i=1}^{n}{\beta_i}{\left\|\bl{P}\bl{h}\otimes\bl{f}-\bl{y}_i\right\|^2}+\sum_{l=1}^{q}\left\|wh^{(l)}\right\|^2+\gamma\left\|\bl{P}\right\|_\mathrm{F} \\
=\sum_{i=1}^{n}{\beta_i}{\left\|\bl{h}\otimes {\bl{P}^\mathrm{T}}\bl{f}-\bl{y}_i\right\|^2}+\sum_{l=1}^{q}\left\|wh^{(l)}\right\|^2+\gamma\left\|\bl{P}\right\|_\mathrm{F},
\end{aligned}
\label{eq:eco}
\end{equation}
where $\beta_i>0$ is  the weight of sample $\bl{x}_i$, $\gamma>0$ is a tuning parameter, $w$ is a spatial penalty to mitigate the drawbacks of the periodic assumption of standard DCF, $q<d$ is the channel number of the projected or compressed filters $\bl{h}$, $\bl{P}$ is a $d\times q$ matrix for performing compression, and $\left\|*\right\|_\mathrm{F}$ is the Frobenius norm. In Eq.\eqref{eq:eco}, by following the linearity of convolution, the factorized convolution alternatively projects the feature vector $\bl{f}$ by matrix-vector product $\bl{P}^\mathrm{T}\bl{f}$, where $\mathrm{T}$ is the transposition operator. Denote $\bl{g}=\bl{P}^\mathrm{T}\bl{f}=\{g^{(1)},...,g^{(q)}\}$, we can follow the similar manner of Eq.\eqref{eq:dcf-cd} and integrate channel distillation into ECO as
\begin{equation}
\begin{aligned}
E(\bl{h},\bl{P},\bl{a}) = \sum_{i=1}^{n}{\beta_i}\left\|\sum_{l=1}^{q}{\alpha_l}{g}^{(l)}\otimes{h}^{(l)}-\bl{y}_i\right\|^2\\
+\frac{q}{\left\|{\bl{a}}\right\|} \sum_{l=1}^{q}{\alpha_l}\left\|wh^{(l)}\right\|^2+\gamma\left\|\bl{P}\right\|_\mathrm{F}, ~~~\text{s.t.}~~\alpha_l \in {\{0,1\}}.
\end{aligned}
\label{eq:eco-cd}
\end{equation}
Then, the loss function in the Fourier domain is derived as,
\begin{equation}
\begin{aligned}
E(\hat{\bl{h}},\bl{P},\bl{a}) = \sum_{i=1}^{n}{\beta_i}\left\|\sum_{l=1}^{q}{\alpha_l}\hat{g}^{(l)}\odot\hat{h}^{(l)}-\hat{\bl{y}}_i\right\|^2\\
+\frac{q}{\left\|{\bl{a}}\right\|} \sum_{l=1}^{q}{\alpha_l}\left\|\hat{w}\otimes\hat{h}^{(l)}\right\|^2+\gamma\left\|\bl{P}\right\|_\mathrm{F}, ~~~\text{s.t.}~~\alpha_l \in {\{0,1\}}.
\end{aligned}
\label{eq:eco-cd-fft}
\end{equation}
From Eq.\eqref{eq:eco-cd} and Eq.\eqref{eq:eco-cd-fft}, channel distillation is applied to perform further compression of feature channels after factorized convolution of ECO, leading to reduced channels, \eg, the number of good channels $c={\left\|{\bl{a}}\right\|}<q$. The optimization is direct by following subsection \ref{subsection:dcf} and ECO~\cite{Danelljan2017CVPR}. During the process of training, good channels $\bl{a}$ are selected from $\bl{g}$ after factorized convolution by following subsection \ref{subsection:dcf} and then loss minimization is performed by following ~\cite{Danelljan2017CVPR} to update $\bl{h}$, which outputs the current solution and loss value $E$. The alternating iteration continues until the maximal iteration number reaches and the final good channels are determined in the iteration with minimal loss value. Therefore, the factorized convolution matrix $\bl{P}$, the filters $\bl{h}$ or $\hat{\bl{h}}$ and the channel selection $\bl{a}$ are jointly optimized in a unified framework.

%In contrast, inspired by the fact that object representation with a pre-trained deep network contains significant redundancy, we integrated channel distillation into a unified energy minimization framework (e.g., DCF) and achieved greater efficiency by optimizing energy to adaptively select good channels. The integrated tracker with DCF can reduce memory and improve speed while having a higher accuracy than its baseline.

To demonstrate the efficacy of channel distillation, we visualize the projected feature maps and the corresponding distilled ones. An example is shown in Fig.~\ref{fig:visualize-eco-cd}, where ECO compresses $608$-channel deep features extracted by the 1st and 5th convolutional layers of VGG-M into $80$ channels and then channel distillation further prunes $16$ channels (marked in red rectangles). In the feature maps, the brighter color indicates larger feature value (more salient activation) and the blue color denotes near-zero value. We can see that most of the pruned feature maps have majority of negligible value, which are hardly contribute to learn correlation filter, especially in the center of the tracked object. In contrast, the remaining feature maps have salient values in the center, which are helpful to learn discriminative correlation filter. In this way, channel distillation can further reduce the negligible channels, saving computation and memory.

\subsection{Tracking Scheme}
Our tracking scheme is shown in Fig.~\ref{fig:framework}, which incorporates channel distillation into the general DCF tracking framework to extract good channels and prune noisy channels. First, based on good channels selected in Step 1, the multi-channel features extracted from the input image are delivered to the channel distillation operator to generate good channel features. Then, cross-correlation is performed with the learned correlation filter in the frequency domain via the fast Fourier transform (FFT). After that, the object position can be predicted according to the response map. Then, the new object is used to update the correlation filter and tracking history, which is used to search and update good channels. In our scheme, we fix the number of good channels for efficacy after tracking one frame. This manner is efficiently used in ECO~\cite{Danelljan2017CVPR}. This channel selection method is efficient due to the small number of training examples.

\section{Experiments} \label{sec:exp}
In this section, we present a comprehensive set of experiments that validate the effectiveness and efficacy of channel distillation. We integrate channel distillation into the standard DCF and state-of-the-art ECO~\cite{Danelljan2017CVPR} formulations and generate our deep trackers, DeepCD and ECO-CD, respectively. Accordingly, we use the deep tracker CF2~\cite{Ma2015ICCV} as the baseline for DeepCD since it follows the standard DCF formulation, while ECO serves as the baseline for ECO-CD. Then, we benchmark the channel-distilled deep trackers on two popular datasets, OTB100~\cite{Wu2015PAMI} and VOT2017~\cite{Kristan2017a}, and we further benchmark against extra thirteen representative and state-of-the-art deep trackers. Moreover, we study the distillation from different channels and small deep models to demonstrate its generalizability and adaptivity. We also analysis effect of different frameworks integrated with channel distillation. These trackers are summarized in Table.~\ref{tab:trackers}. In the experiments, channel distillation is carried out in the first two frames.

%More experimental results are provided in \textbf{supplementary Material}.
\begin{table*}[t]
\caption{The baseline and evaluated deep trackers. The baselines are marked with underline.}
%\vspace{5.0pt}
\centering{
%\scriptsize
\begin{tabular}{@{}llccl@{}}
\hline
\textbf{Tracker}                    &\textbf{Category}   &\textbf{Model of Features} &\textbf{\#Channels} &\textbf{Published}  \\
\hline
\underline{CF2}\cite{Ma2015ICCV}    &Weighting   &VGG-19                  &1,280                &2015 (ICCV)  \\
DeepSRDCF\cite{Danelljan2015ICCVW}  &Weighting   &VGG-M                   &96                   &2015 (ICCVW)  \\
HDT\cite{Qi2016CVPR}                &Weighting   &VGG-19                  &3,072                &2016 (CVPR)   \\
ADNet\cite{Yun2017CVPR}             &Weighting   &Learned                 &864                  &2017 (CVPR)   \\
DSLT\cite{DSLT}                     &Weighting   &VGG-16                  &1,024                &2018 (ECCV)   \\
TM$^3$-deep\cite{Liu2018TIP}        &Weighting   &VGG-16                  &4,096                &2018 (TIP)   \\
SiamFC\cite{Bertinetto2016ECCVW}    &Learning    &Learned                 &256                  &2016 (ECCVW)   \\
CFNet\cite{Valmadre2017CVPR}        &Learning    &Learned                 &256                  &2017 (CVPR)   \\
DCFNet\cite{DCFNet}                 &Learning    &Learned                 &32                   &2017 (Arvix)   \\
RASNet\cite{RASNet}                 &Learning   &Learned                  &256                  &2018 (CVPR)   \\
C-COT\cite{Danelljan2016ECCV}       &Compression &VGG-M                   &611                  &2016 (ECCV)   \\
TRACA\cite{Choi2018CVPR}            &Compression &VGG-M                   &256                  &2018 (CVPR)   \\
CFCF\cite{CFCF}                     &Compression    &VGG-M                &611                  &2018 (TIP)   \\
CRT\cite{CRT}                       &Compression    &VGG-16               &64                   &2018 (TIP)   \\
\underline{ECO}\cite{Danelljan2017CVPR}  &Compression &VGG-M              &80                   &2017 (CVPR)   \\
%CBCW\cite{CBCW}                    &-----       &HOG+CN                  &41                   &2018 (TIP)   \\
\hline
\end{tabular}}
\label{tab:trackers}
\end{table*}
\begin{figure}[t]
  \centering{\includegraphics[width=0.49\linewidth]{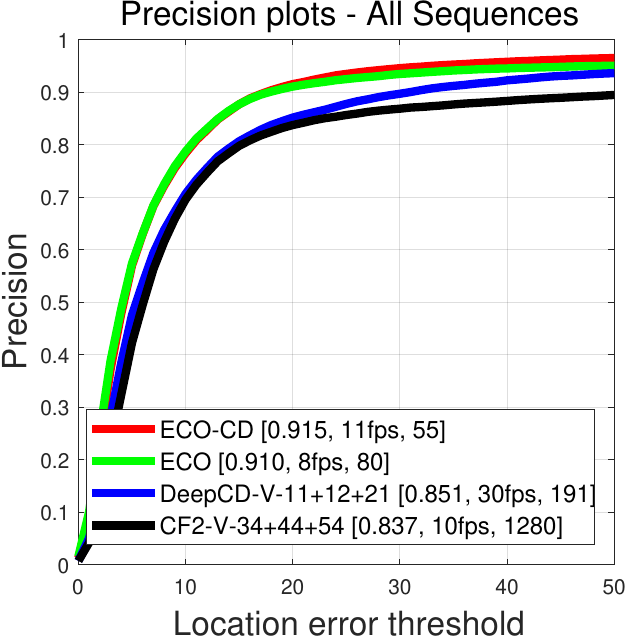}}
  \centering{\includegraphics[width=0.49\linewidth]{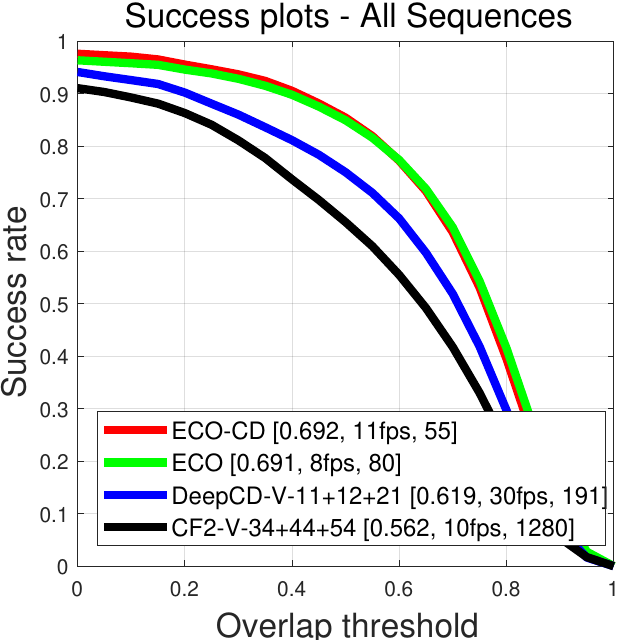}}
  \caption{Baseline comparison on OTB100. The measure, speed and the number of average feature channels used for each tracker are shown in the legend.}
\label{fig:res-baseline-otb100}
\end{figure}

\begin{figure*}[t]
  \centering{\includegraphics[width=1.0\linewidth]{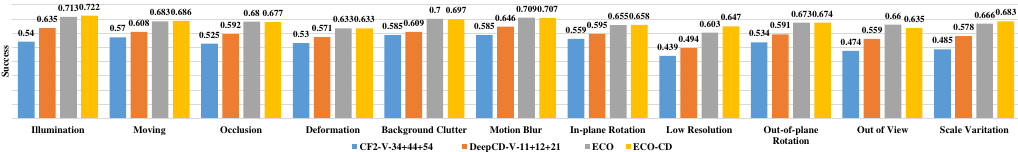}}
  \caption{Baseline comparison under various circumstances on OTB100}
\label{fig:res-baseline-otb100-condition}
\end{figure*}

\begin{figure*}[t]
  \centering{\includegraphics[width=1.0\linewidth]{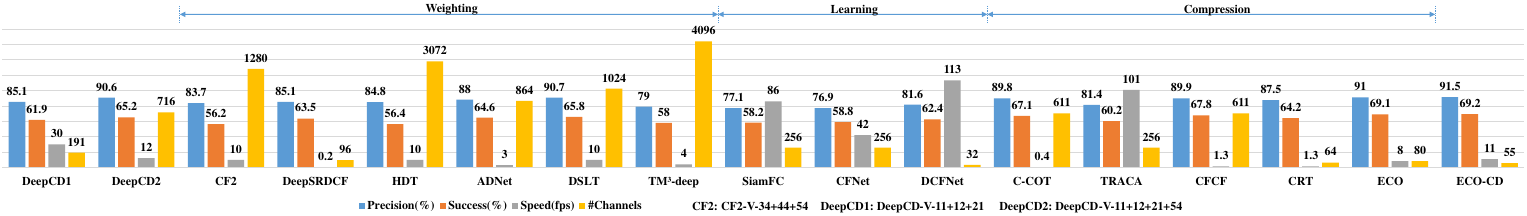}}
  \caption{State-of-the-art comparison on OTB100.}
\label{fig:res-state-otb100}
\end{figure*}

\subsection{Baseline Trackers and Evaluation}
\myPara{Baseline Trackers}. We select two representative deep trackers, CF2~\cite{Ma2015ICCV} and ECO~\cite{Danelljan2017CVPR}, and apply channel distillation to the baselines, calling the good channel versions DeepCD-$\bm{N}$-$\bm{L}$ and ECO-CD, respectively. Here, $\bm{N}$ is the deep network (\eg, ``V'' for VGG-19) used and $\bm{L}$ is the layer set for candidate channel selection (\eg, ``34+44+54'' for conv34, conv44 and conv54). Similarly, CF2 is represented as CF2-$\bm{N}$-$\bm{L}$ for the sake of simplicity. CF2 uses a pre-trained VGG-19 model~\cite{Simonyan2015ICLR} to extract multi-channel deep features from different layers for sperate correlation filtering and then fuses the responses together to form the final output. ECO compresses the features from the pre-trained VGG-M model to achieve an optimized tracker. CF2-V-34+44+54 used in~\cite{Ma2015ICCV} has $1,280$ feature channels. In contrast, DeepCD-V-$\bm{L}$ selects and combines good channels from some layers $\bm{L}$ of VGG-19 model. VGG-19 and VGG-M are all trained for object classification on ImageNet and achieve a top-5 accuracy of $90.1\%$ and $84.2\%$, respectively. {We use DeepCD-V-11+12+21 which distills from early layers as our tracker for baseline comparison and also study other trackers distilling from various layers.}

\myPara{Evaluation on OTB100}. All experiments are evaluated using two measures \cite{Wu2013CVPR,Wu2015PAMI}: precision and success. Precision measures the center error between the ground truth bounding box and tracker bounding box, while success is measured as their intersection-over-union~(IoU). In the precision or success plot, the maximum allowed center error in pixel distance or the required overlap is varied along the x-axis, and the percentages of the correctly predicted tracker bounding boxes per threshold are plotted on the y-axis. According to \cite{Wu2013CVPR}, trackers are ranked by the common threshold of $20$ pixels for precision and area under the curve~(AUC) for success. All the results are generated by OTB-Toolkit \cite{Wu2015PAMI}. %Similar evaluation method is used on TC128~\cite{TC128}.
In the experimental results, the measure, speed and the number of feature channels used in each tracker are shown.

\myPara{Evaluation on VOT2017}. Following the general measure method in \cite{Kristan2017a}, we apply overall expected average overlap (EAO) scores to evaluate overall tracking performance in both accuracy and robustness. Larger EAO scores represent better tracking performance. All the results are generated with VOT-Toolkit \cite{Kristan2017a}.

\begin{table}[h]
\centering
%\scriptsize
\caption{Baseline comparison on VOT2017. The larger EAO score represents higher performance.}
\label{tab:res-baseline-vot2017}
\begin{tabular}{lcc}
\hline
\textbf{Tracker}     &\textbf{EAO}      &\textbf{\#Average Channels}  \\
\hline
CF2-V-11+12+21       &0.116             &256                 \\
DeepCD-V-11+12+21    &\textbf{0.127}    &\textbf{195}         \\
%\hline
ECO                  &0.281             &80                 \\
ECO-CD               &\textbf{0.298}    &\textbf{53}         \\
\hline
\end{tabular}
\end{table}

\begin{figure}[t]
   \centering{\includegraphics[width=1.0\linewidth]{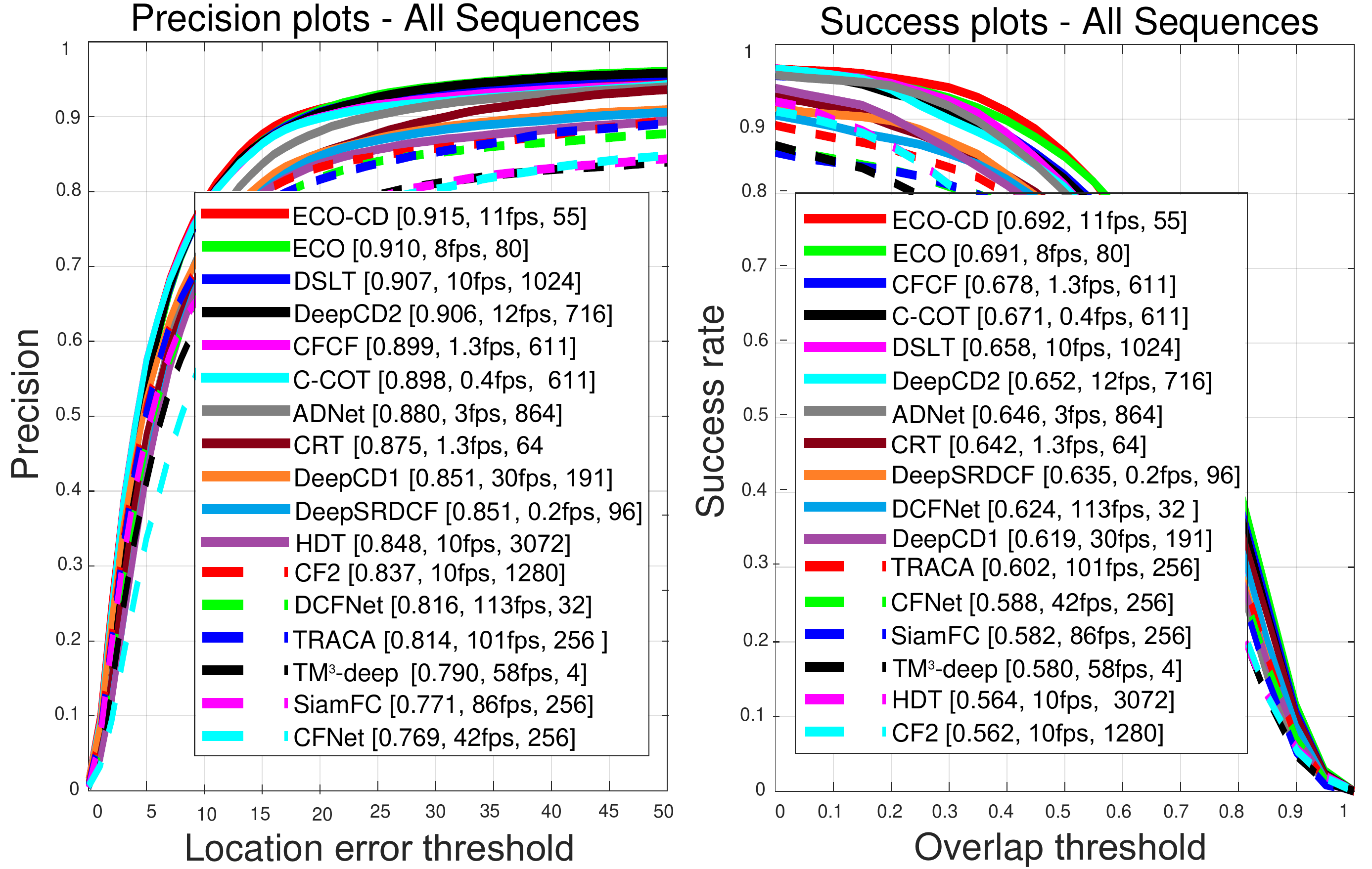}}
  \caption{Precision and success plots of state-of-the-art comparison on OTB100.}
\label{fig:otb100-ps}
\end{figure}

\subsection{Baseline Comparison}
To validate the effectiveness and generalizability of our channel distillation framework for improving tracking performance, we examine the results of the representative baseline trackers using its good channel version on two popular benchmarks (OTB100 and VOT2017) which contain various videos with diverse objects and different performance measures.

\myPara{OTB100}. The results of the baseline trackers and their good channel counterparts on OTB100 are shown in Fig.~\ref{fig:res-baseline-otb100}. We find that the good channel versions outperform the baselines in both precision and success rate, with a $5.7$\% improvement (from $0.562$ to $0.619$) in success rate for DeepCD-V-11+12+21. {When distilling from more and deeper layers, the performance can be further improved (see Fig.~\ref{fig:res-deep-cd}), e.g., 0.652@success achieved by DeepCD-V-11+12+21+54.} For ECO-CD, we only distill the deep feature channels from $1$st and $5$th layers in VGG-M, leading to an improvement of $0.5$\% (from $0.910$ to $0.915$) in precision and a more reduced channel number of $55$ than the projected $80$ channels. Therefore, channel distillation can increase the processing speed remarkably. In addition, the comparable improvement can be found under different circumstances, as shown in Fig.~\ref{fig:res-baseline-otb100-condition}. { Similar results are also achieved on TC128 benchmark~\cite{TC128}. For example, DeepCD-V-11+12+21 achieves $0.536$@success at $34$ fps when costing $191$ channels, and CF2-V-34+44+54 delivers $0.495$@success at $12$ fps with $1,280$ channels.} These results imply that channel distillation is a general framework for visual tracking in different videos that can effectively extract the essential channels and reduce the influence of the noisy channels.

\myPara{VOT2017}.~To verify the generalizability of our framework, we further investigate it on VOT2017, a very challenging tracking benchmark. In this set of experiments, CF2 and its good channel version use the features from three early layers (conv11, conv12, and conv21) of VGG-19, then form CF2-V-11+12+21 and DeepCD-V-11+12+21, respectively. ECO and ECO-CD all use VGG-M to extract channel features. Table~\ref{tab:res-baseline-vot2017} lists their EAO scores and the average number of feature channels used. It can clearly be seen that DeepCD-V-11+12+21 achieves a larger EAO score than its baseline. Moreover, the average number of feature channels is reduced to $192$ from $256$, suggesting that channel distillation could improve tracking performance under different measures. ECO-CD also gives a higher EAO score than ECO while costing less feature channels ($53$ vs. $80$). Due to the reduction of feature channels, ECO-CD gives a faster speed, such as 11 fps against 8 fps in ECO. { That is to say, even ECO has reduced filters to deliver high accuracy, our channel distillation still can further reduce feature channels to achieve slightly higher tracking accuracy and considerably faster running speed.}

\subsection{State-of-the-art Comparison}
We next perform comparison with fifteen state-of-the-art trackers on OTB100: two baselines (CF2 and ECO), five weighting-based trackers (DeepSRDCF~\cite{Danelljan2015ICCVW}, HDT~\cite{Qi2016CVPR}, ADNet~\cite{Yun2017CVPR}, DSLT~\cite{DSLT} and TM$^3$-deep~\cite{Liu2018TIP}), four learning-based trackers (SiamFC~\cite{Bertinetto2016ECCVW}, CFNet~\cite{Valmadre2017CVPR}, DCFNet~\cite{DCFNet} and RASNet~\cite{RASNet}), and four compression-based trackers (C-COT~\cite{Danelljan2016ECCV}, TRACA~\cite{Choi2018CVPR}, CFCF~\cite{CFCF} and CRT~\cite{CRT}).
The results are shown in Fig.~\ref{fig:res-state-otb100} { as well as Fig.~\ref{fig:otb100-ps} for better visualization}, where DeepCD-V-11+12+21 and DeepCD-V-11+12+21+54 distill good channels from different convolutional layers. The larger tracker DeepCD-V-11+12+21+54 outperforms twelve trackers and is comparable to DSLT and ECO, whilst also being faster with respect to precision. DeepCD-V-11+12+21+54 also surpasses ten trackers with respect to success, and gives higher precision and success than all the weighting-based trackers.
Moreover, the smaller tracker DeepCD-V-11+12+21 still achieves good performance, even distilling from early layers. When distilling on ECO, the resulting ECO-CD surpasses all the trackers with respect to both precision and success,whilst being faster than ECO.

\begin{figure}[t]
  \centering{\includegraphics[width=1.0\linewidth]{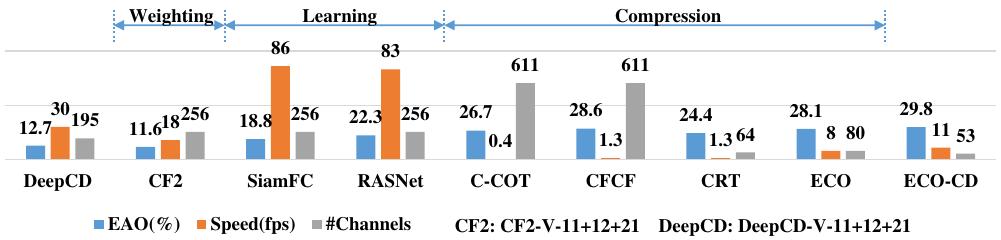}}
  \caption{State-of-the-art comparison on VOT2017.}
\label{fig:res-state-vot2017}
\end{figure}

On VOT2017, we perform comparison with seven state-of-the-art trackers: two weighting-based trackers (DeepSRDCF~\cite{Danelljan2015ICCVW} and RASNet~\cite{RASNet}), three learning-based trackers (SiamFC~\cite{Bertinetto2016ECCVW}, CRT~\cite{CRT} and CFCF~\cite{CFCF}) and two compression-based trackers (C-COT~\cite{Danelljan2016ECCV} and ECO~\cite{Danelljan2017CVPR}). Fig.~\ref{fig:res-state-vot2017} shows the results. The distilled on standard DCF tracker, DeepCD-V-11+12+21, gives a higher EAO value than its baseline CF2-V-11+12+21 and cost much less channels. ECO-CD which distills on more advanced ECO framework surpasses all the trackers with respect to EAO, implying that channel distillation can consistently improve the tracking performance when incorporating into a DCF framework.

\subsection{Efficacy Analysis}
{\color{black}We next examine speed and memory gains, and verify that channel distillation can improve tracking performance on different convolutional layers.} We study two baseline trackers: DeepCD-V-11+12+21 and CF2-V-34+44+54, as shown in Fig.~\ref{fig:res-baseline-otb100}. The tracking speed mainly arises from feature computation with deep network inference and model matching with correlation filtering, while the memory consumption greatly depends on the number of channels in storing feature maps and tracking model.

First, DeepCD-V-11+12+21 performs foreword inference in earlier three convolutional layers, costing much less feature computation than CF2-V-34+44+54 which employs later three convolutional layers. In addition, channel distillation significantly reduces the complexity of cross-correlation for model computation and matching. Therefore, channel distillation achieves impressive speed gains.

When DeepCD-V-11+12+21 distills features from three shallow layers (conv11, conv12, and conv21) of VGG-19, the average numbers of good channels in these three layers are reduced to $28.4$, $27.7$ and $44.3$ from $64$, $64$ and $128$, respectively. {Moreover, the less channels in the first two layers favours larger memory reduction due to bigger size of their feature maps. DeepCD-V-11+12+21 therefore takes a less memory cost of 12.2MB. In contrast, CF2-V-34+44+54 represents features with three deeper layers, resulting in much heavier memory cost since the feature maps from the early to these deeper layers all need to store. This implies the effective reduction by our approach in memory cost.}  This reduction in feature channels greatly benefits memory and speed when running the tracker and could facilitate real-world applications.

{ We further compare with two Siamese-based trackers that have faster speed: SiamFC~\cite{Bertinetto2016ECCVW} and RASNet~\cite{RASNet}. They all adopt a pretrained network for object representation and cost 9.4MB and 18.6MB in memory, respectively. Our DeepCD-V-11+12+21 takes a comparative memory cost of 12.2MB.}

\begin{figure}[t]
  \centering{\includegraphics[width=0.49\linewidth]{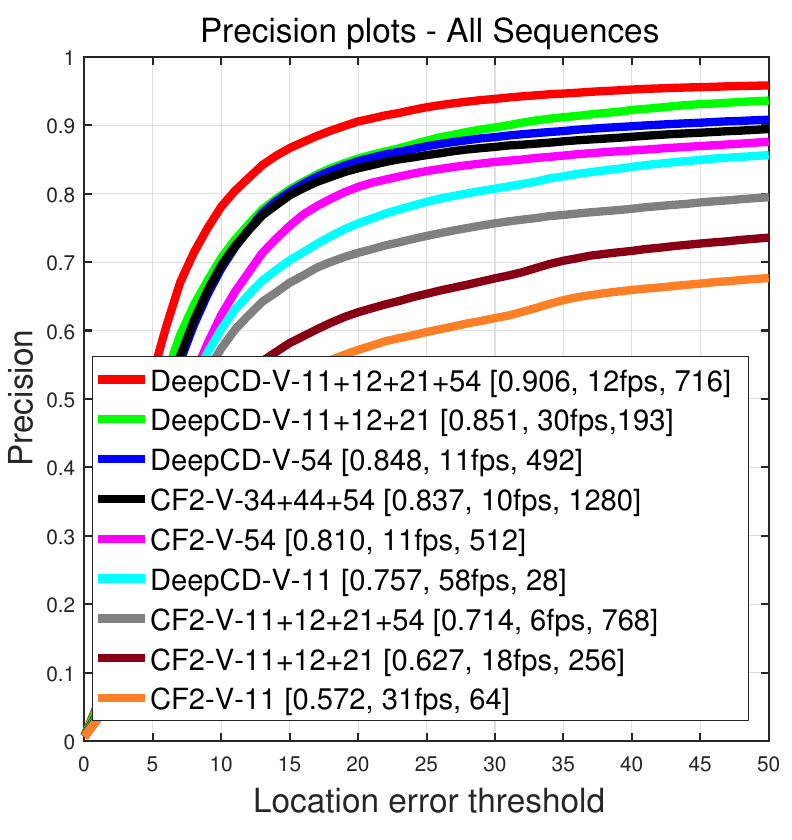}}
  \centering{\includegraphics[width=0.49\linewidth]{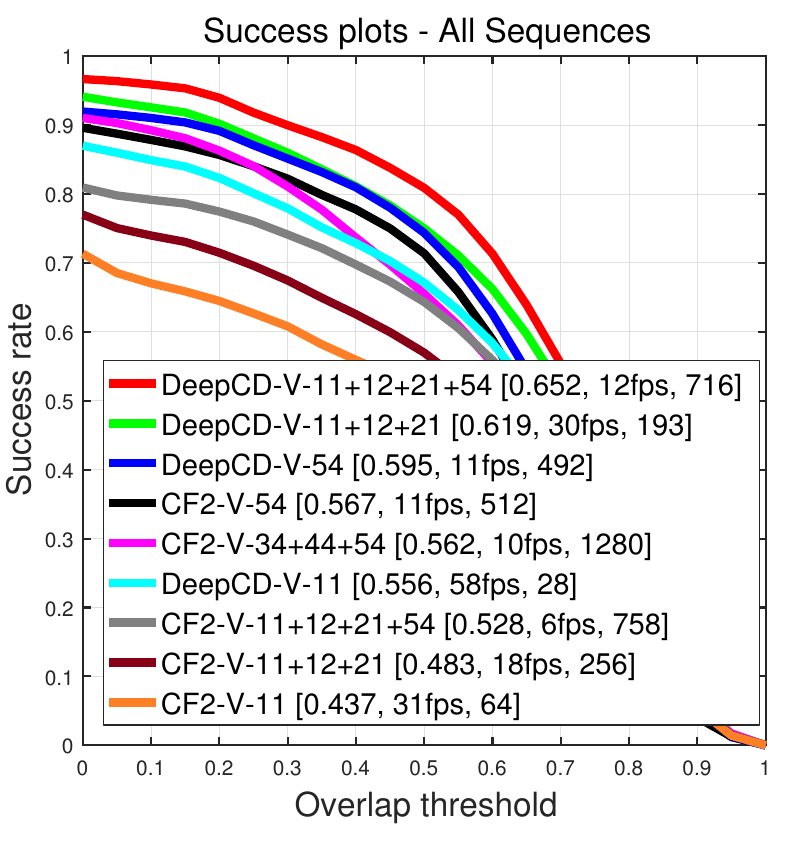}}
  \caption{Tracking results on OTB100 with distillation on different channels.}
\label{fig:res-deep-cd}
\end{figure}

\subsection{Distillation from Different Channels}
Despite high accuracy, deep trackers are usually computationally expensive and consume large amounts of memory due to feature representation from certain deep layers~\cite{Ma2015ICCV,Qi2016CVPR,Valmadre2017CVPR}. This makes their deployment challenging. To provide helpful guidance for their practical deployment, we investigate the effect of channel distillation on tracking performance and speed when selecting different deep feature channels.

Fig.~\ref{fig:res-deep-cd} shows some results on OTB100. We note the following: channel distillation always improves over baseline, with both channel reduction and performance improvements more obvious when distilling on early layers since they contain more noisy channels. Selecting features from more layers improves accuracy but degrades speed. Thus, feature channels should be selected from certain layers according to deployment conditions. For example, one-layer DeepCD-V-11 retains competitive precision (0.757) and speed (58 fps) whilst only requiring 7KB, making it highly suitable for resource-limited deployment. If more storage is available, DeepCD-V-11+12+21 selecting from three early layers of VGG-19 has a higher precision (0.851) and is still very fast (30 fps) whilst being $500\times$ smaller (0.14MB) than its baseline (71MB). When the channels are distilled from more final layers, the precision reaches 0.906, which is higher than C-COT~\cite{Danelljan2016ECCV} (0.898) and comparable to ECO~\cite{Danelljan2017CVPR} (0.910).

\begin{figure}[t]
  \centering{\includegraphics[width=0.49\linewidth]{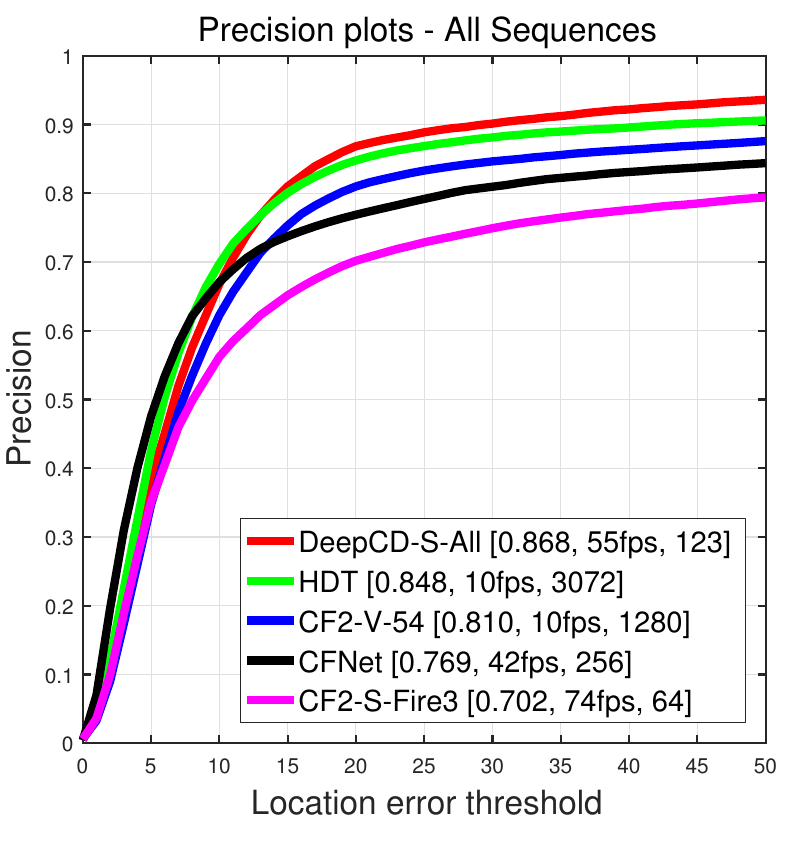}}
  \centering{\includegraphics[width=0.49\linewidth]{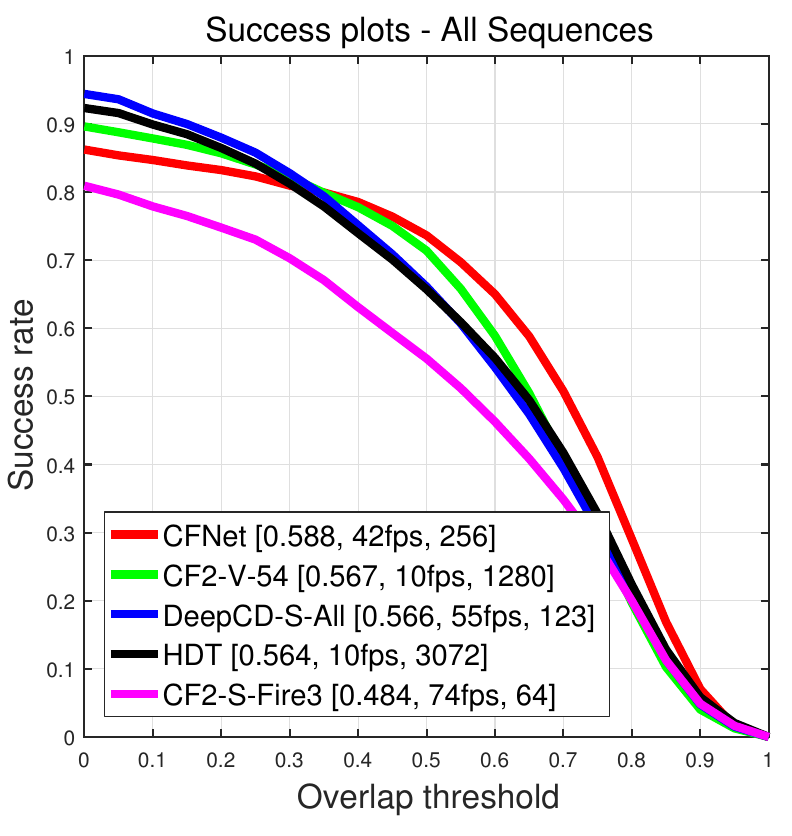}}
  \caption{Tracking results on OTB100 with distillation on different deep networks.  Weighting-based tracker HDT and learning-based tracker CFNet are provided for comparison.}
\label{fig:res-distil-small-model}
\end{figure}

\subsection{Distillation from Small Deep Networks}
To further validate the effectiveness and generalizability of channel distillation, we investigate its performance when distillation is performed on small deep networks. We use SqueezeNet~\cite{Iandola2017ICLR} for example. SqueezeNet is a small network for object classification with $4.8$M parameters and gives a top-5 accuracy of $80.3\%$ on ImageNet. It consists of one convolutional layer, eight fire modules, and one final convolutional layer. A fire module begins with a squeeze convolutional layer with $1\times 1$ filters and ends with an expand layer including $1\times 1$ and $3\times 3$ convolution filters. We distill all the layers before Softmax for feature extraction, resulting in a deep tracker, DeepCD-S-All.
As shown in Fig.\ref{fig:res-distil-small-model}, DeepCD-S-All achieves the highest precision and speed whilst only needing $123$ channels. DeepCD-S-All also achieves the top success rate when the overlap threshold is lower than $0.32$, although the total success rate is a little lower than CFNet.

Channel distillation also shows better generalizability. For example, by using channel distillation, DeepCD-S-All achieves comparable performance to DeepCD-V-54 ($2\%\uparrow$ in precision and $2.9\%\downarrow$ in success). However, directly using deep features from small and big networks without distillation, CF2-S-Fire3 which uses fire3 module layer of SqueezeNet has a sharp performance drop to CF2-V-54 ($10.8\%\downarrow$ in precision and $8.3\%\downarrow$ in success). Despite higher accuracy achieved by using a more powerful deep network (\eg, VGG-19), CF2-V-54 is computationally more expensive and consumes much larger amounts of memory. It reveals that a deep network without distillation has poor generalization to represent a specific moving object, and the resulting deep trackers greatly rely on the pre-trained deep networks.
In contrast, channel distillation can adaptively distill the correct knowledge from general knowledge learned from massive data, and transfer it to track a specific object on the fly.

\subsection{Distillation on Different Frameworks}
We also investigate the results when distilling on three different DCF frameworks: a standard DCF in CF2 with deep features from VGG-19, a recent advanced DCF termed CACF~\cite{Mueller2017CVPR} with HOG features and SRDCF of ECO with deep features from VGG-M. To this end, we further incorporated channel distillation into CACF, resulting in CACF-CD. As a result, on OTB100, the channel-distilled versions DeepCD-V-11+12+21, CACF-CD and ECO-CD give 0.619@success at 30 fps, 0.551@success at 120 fps, and 0.692@success at 11 fps, respectively. The performance is consistently better than their baselines CF2-V-11+12+21 (0.483@success at 18 fps), CACF (0.511@success and at fps) and ECO (0.691@success at 8 fps). It implies that the robustness of the framework applied and the feature extractor is critical for improving tracking performance. For example, CF2 which uses standard DCF and deep features performs worse than ECO with advanced SRDCF framework, but is better than CACF with hand-crafted HOG features. We suspect that adaptively selecting better features and incorporating into more advanced framework allow a visual tracker to track specified objects efficiently and effectively. { What's more, we conduct CACF-CD on TC128 benchmark~\cite{TC128} to  verify the generalizability of the proposed approach and achieve 0.584@success at 144 fps, outperforming the baseline CACF (0.509@sussess at 116 fps).}

{Beyond DCF frameworks, we further check the distillation on MDNet~\cite{Nam2016CVPR} that is not a DCF-based tracker. MDNet first obtains a generic feature representation by pretraining a CNN that is composed of 3 convolutional layers with 96, 256 and 512 channels,  two 512-way fully-connected layers and multiple branches of domain-specified layers, and then performs tracking by online updating the two fully-connected layers and a domain-specified binary classification layer. We conduct the experiment on OTB100 and compare the channel-distilled version MDNet-CD with its baseline that has achieved 0.909@precision and 0.678@success. In our experiment, we distill channels on the 3rd convolutional layer and online update two following fully-connected layers and domain-specified classification layer. Finally, MDNet-CD delivers 0.912@precision and 0.692@success when compressing original 512 channels to 239 channels, implying the general effectiveness of channel distillation.
}

\section{Conclusions}
Many DCF trackers with deep features have been proposed, showing good performance for visual tracking. The multi-channel deep features used are usually fully fixed to represent diverse objects. This fixed-channel setting is suboptimal, degrades tracking performance, and is slow. However, it is possible to select good channels that are more effective for the tracked object. This paper studies the influence of channel pruning and proposes a generic scheme to select good channels. We show that channel selectivity exists, good channels can be found to track objects, and that the DCF framework improves tracking accuracy and speed as well as generalizability when integrated with channel distillation. We believe that this finding and channel distillation approach will facilitate the development of efficient visual tracking in real-world applications. Future work will include joint distillation and compression for deep trackers and other applications of channel distillation.

\myPara{Acknowledgement}. This work was partially supported by grants from the National Natural Science Foundation of China (61772513), the project from Beijing Municipal Science and Technology Commission (Z191100007119002), National Key Research and Development Plan (2016YFC0801005), and the Open Foundation Project of Robot Technology Used for Special Environment Key Laboratory of Sichuan Province in China (16kftk01). Shiming Ge is also supported by the Open Projects Program of National Laboratory of Pattern Recognition, and the Youth Innovation Promotion Association, Chinese Academy of Sciences. Shiming Ge is the corresponding author.

\bibliographystyle{IEEEtran}
\bibliography{TIP-CD}


%\newpage


%\vfill

\end{document}